\documentclass[peerreview]{IEEEtran}
\IEEEoverridecommandlockouts
\usepackage[table]{xcolor}
\usepackage{graphicx}
\usepackage{float}
\usepackage{verbatim}
\usepackage{xcolor}
\usepackage{multirow}

\title{A Survey on Impact of Transient Faults on BNN Inference Accelerators}

\author{Navid Khoshavi$^{1,2}$, Connor Broyles$^2$, Yu Bi$^{3}$
\\$^1$Department of Computer Science, Florida Polytechnic University
\\ $^2$Department of Electrical and Computer Engineering, Florida Polytechnic University
\\$^3$Department of Electrical and Computer Engineering, University of Rhode Island}
\begin{document}
\maketitle

% check this link
%http://www.esa.int/Our_Activities/Space_Engineering_Technology/ESA_team_blasts_Intel_s_new_AI_chip_with_radiation_at_CERN
\begin{abstract}
Over past years, the philosophy for designing the artificial intelligence algorithms has significantly shifted towards  automatically extracting the composable systems from massive data volumes. This paradigm shift has been expedited by the big data booming which enables us to easily access and analyze the highly large data sets. The most well-known class of big data analysis techniques is called \textit{deep learning}. These models require significant computation power and extremely high memory accesses which necessitate the design of novel approaches to reduce the memory access and improve power efficiency while taking into account the development of domain-specific hardware accelerators to support the current and future data sizes and model structures. The current trends for designing application-specific integrated circuits \textit{barely consider the essential requirement for maintaining the complex neural network computation  to be resilient in the presence of soft errors}. The soft errors might strike either memory storage or combinational logic in the hardware accelerator that can affect the architectural behavior such that the precision of the results fall behind the minimum allowable correctness. In this study, we demonstrate that the impact of soft errors on a customized deep learning algorithm called Binarized Neural Network might cause drastic image misclassification. Our experimental results show that the accuracy of image classifier can drastically drop by 76.70\% and 19.25\% in lfcW1A1 and cnvW1A1 networks, respectively across CIFAR-10 and MNIST datasets during the fault injection for the worst-case scenarios.

%Our experimental results shows that the accuracy of image classifier in cnvW1A1 network can significantly drop from 80.5\% to 72.0\%, on average across CIFAR-10 dataset during the fault injection.

%Due to the significance of side channel attacks, the challenges and overheads of these attack vectors are described through the course of this study.

%Deep learning presents an important field for the development of domain-specific computer architectures. With the increasing challenges faced in maintaining the continuity of Moore's Law and the successes of deep learning algorithms on traditionally challenging categories of learning tasks, the effort required to develop domain specific architectures for deep learning is attracting significant attention.

\end{abstract}

\begin{IEEEkeywords} 
\textit{Fault Injection, Deep Neural Network Accelerator, Machine Learning, Soft Error}
\end{IEEEkeywords} 

%\begin{figure}[h!]
%\caption{Memory Hierarchy}
%\centering
%\includegraphics[width=0.48\textwidth]{Fig/Backpropagation.png}
%\end{figure}

\section{Introduction}
\label{sec:introduction}
Deep learning has had a long and rich history. The increasing datasets have spurred the rapid development of machine learning methods. Deep learning Neural Networks (DNNs), which mimics humans' brain activities, have been surging as an effective and efficient method to solve a variety of existing big data problems. Computer vision research, e.g. images and videos, arguably makes the largest impact for today's DNN study, given that humans are heavily reliant on sights for the information \cite{abdel2014convolutional}. The popular self-driving research is one of applications that particularly leverages the advance of computer vision research. Besides, DNNs have also provided the avenue for the other fields of interest, such as natural language, speech recognition, robotics, biomedical applications and many more.

The superior accuracy of DNNs, however, comes at the cost of high computational complexity and considerable off-chip memory accesses. Due to the demands of thousands of parallel computing, Graphic Processing Units (GPUs) have been mainly adopted for high-performance hardware platform, while Field-Programmable Gate Arrays (FPGA) are employed for less intensive computing platform, such as in-car computers. Besides, along with the progress of DNN algorithms, scholars have been largely working on developing the specialized hardware designs to accelerate the \textit{inferencing} process of neural networks. To be specific, the two primary phases to build and utilize the DNN model are training and inference modes. In order to train the DNN model, a gigantic dataset and tons of computation power are required which make this process significantly slow. Nevertheless, once the training phase is completed, the inference model can quickly perform the prediction on the given input as long as the re-training is not required. For instance, Google has deployed its in-house hardware accelerator, so called Tensor Processing Unit (TPU), in datacenter for speeding up DNN applications. 

%MS
It is worth noting that a parallel approach is to reduce the data size while maintaining the performance. \cite{Sedghi2020rep} is a preeminent work studying the problem of representative selection along this line. Not to compromise on the training power, the authors propose to capture the global structure of huge datasets through non-linear manifolds, hence, offering strong generalization capabilities. The method has shown improved performance over the state-of-the-art subset selection methods, while bringing about substantial speed-
up.

\begin{figure}[t!]
\centering
\includegraphics[width=0.48\textwidth]{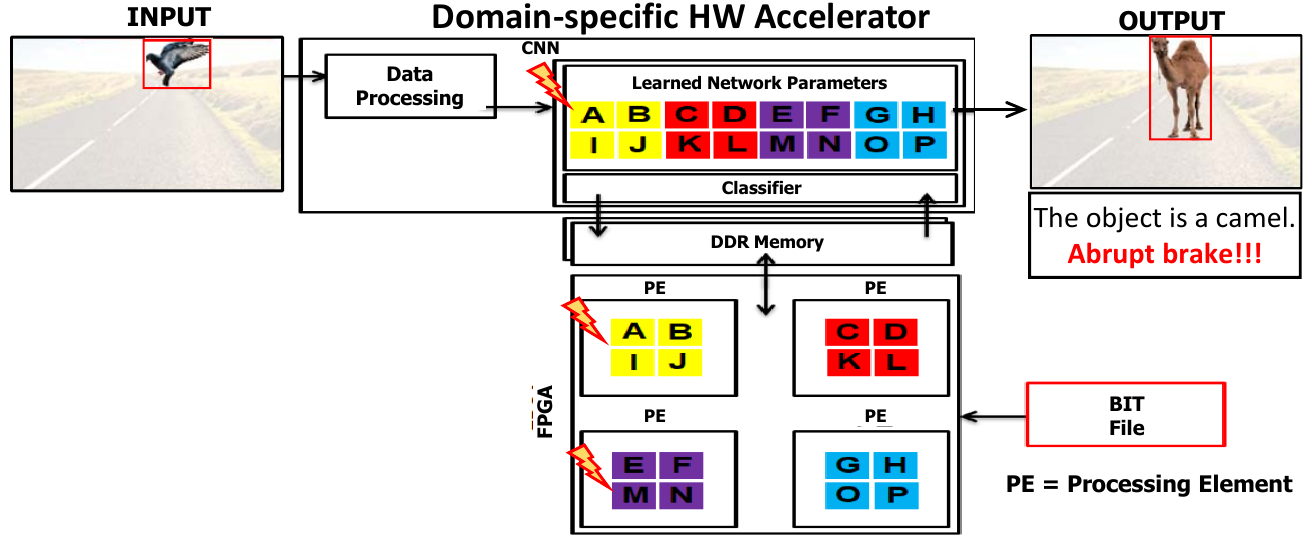}
\vspace*{-0.3cm}
\caption{Soft error impacts on different locations in a DNN accelerator might result in image misclassification. This might cause the self-driving car to make a hard brake if images are used to define the driving actions (adopted from \cite{lacey2016deep}).} 
\label{soft_error}
\end{figure}

Another effort that falls in the category of inference accelerators is Binarized Neural Network (BNN) inference accelerator \cite{courbariaux2016binarized}. BNN compresses the network information in a reduced memory footprint through representing all input activations, weights and output activations with 1-bit and 2-bit. Even though this data representation has significantly reduced the costly off-chip memory accesses, but it also has increased the impact of transient faults on the results. In particular, condensing a 32-bit floating point number which was originally used to demonstrate weights and activations to few bits comes with the risk of loosing the whole tensor information if the representative BNN bit set flips due to soft errors, which is triggered by high-energy particles striking transistors, can cause outlier contamination, and malfunctions such as flip bit value in sequential logic and glitch in combinational logic \cite{khoshavi2016bit}. %MS
Machine learning models have been shown to be drastically distorted in presence of outliers \cite{Sedghi2019GCP,Sedghi2018CP}.  Needless to say that BNN inference accelerator is expected to remain functional for a significantly long period. Thus, the accumulated soft errors can gradually downgrade the output accuracy in the accelerator. 
In particular, if the soft error causes the data corruption that will be reused later in the dataflows of BNN accelerator, the contaminated data will pollute any remaining steps in computations.% which results in drastic accuracy degradation in the image classification. 
$~$For instance, as illustrated in Fig. \ref{soft_error}, the soft error can impact different locations in a BNN accelerator which is employed in a self-driving car. This incident might result in image misclassification during the safety-critical mission and might end up with a potentially dangerous consequences. In this paper, we will present a comprehensive study on the impact of soft errors on BNN accelerators. Specifically, our contributions in this study are as follows:

\begin{itemize}
    \item We investigate the behavior of two well-known categories of soft errors, Single-Event Upset (SEU) and Multi-Bit  Upset (MBU), across the time and space on the combinational logic and memory storage in BNN accelerator.
    
    \item We examine the effect of soft errors on the different network topologies and data types used to represent the weights, activations, and different layers. This part of our study identifies the most vulnerable parameters against soft errors in a BNN inference accelerator. 
    
    \item We propose a fault injection scenario on a modified version of FINN framework \cite{umuroglu2017finn}. Our approach is significantly more accurate compared to the existing software-level fault injections such as \cite{li2017understanding} while targeting the architectural BNN accelerator. We inject the faults uniformly across time and space on a Xilinx  Zynq-7000 ARM/FPGA SoC while the board is executing the classification workload.
    
    \item We demonstrate that the classification accuracy in BNN accelerator can  drastically drop by \textbf{76.70\%} and \textbf{19.25\%} in \textit{lfcW1A1} and \textit{cnvW1A1} networks, respectively in the presence of soft errors for the worst-case scenarios.   
\end{itemize}

The remainder of the paper is organized as follows. Sec. II presents
the preliminaries of this work. In Sec. III, the experimental results are presented.  Finally, the Section IV concludes the paper.

% is the fault injected on data layout (weights, configuration, etc) -> sequential logic or it also is injected on the combinational logic.

% what does the targeted memory for fault injection represents. is it CLB in fpga? LUT?

% how the fault has been injected? Does in follow guassian distribution?

\section{PRELIMINARIES}
\subsection{Neural/Deep Neural/Convolutional Networks}
\begin{figure}[t!]
\centering
\includegraphics[width=0.35\textwidth]{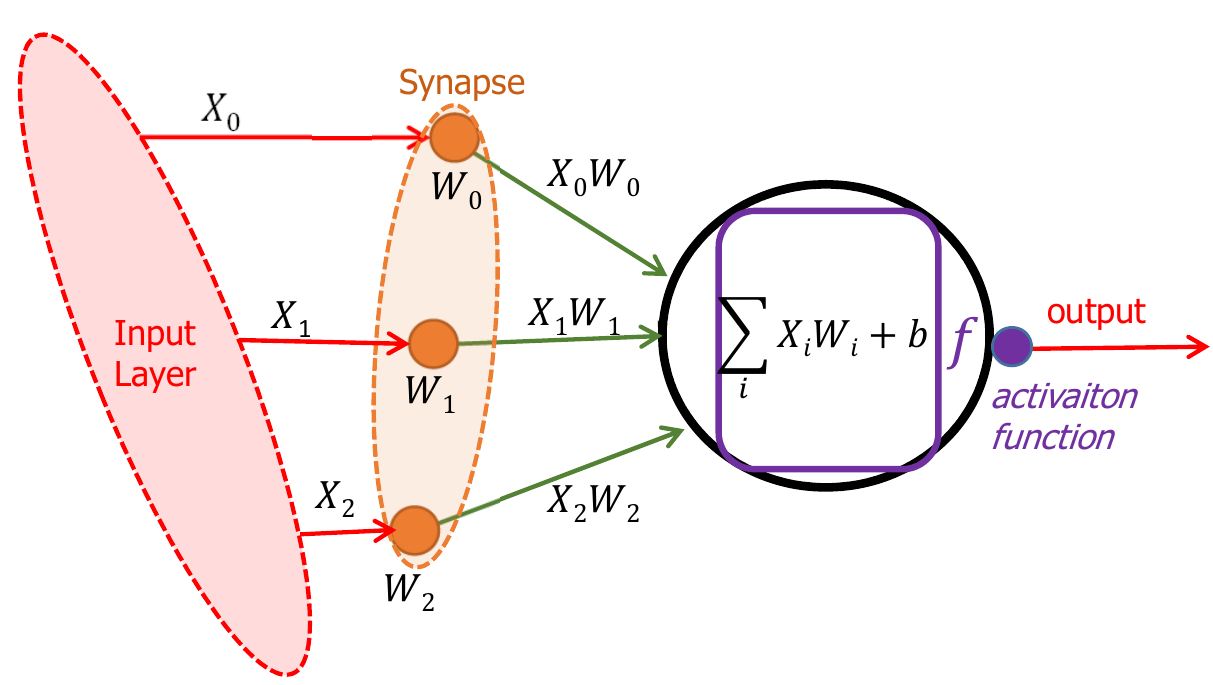}
\vspace*{-0.3cm}
\caption{Human brain-inspired neuron model}
\label{NN_model}
\end{figure}

A Neural Network (NN) is a computing system inspired by the human brain that is composed of three layers: an input layer, a hidden layer, and an output layer.  Each of these layers are composed of neurons while the connections between these layers are considered synapses.  A neuron in a NN, as illustrated in Fig. \ref{NN_model}, is effectively a function that takes in the outputs of all of the neurons in the previous layer and produces an activation, though neurons in the input layer provide a value inherently. 
%This function is composed of 3 parts: (1) the weighted sum of activations from the previous layer, (2) the bias of that particular neuron for shifting the activation function to the left or right, (3) the non-linear activation function that operates as a threshold for certain values. The synapses represent the weights that are responsible for the value scaling of the activations before they are used by the following neuron. The \textit{activation function} can be categorized to two groups: 1) linear activation functions, and 2) non-linear activation functions. Since the linear activation cannot confine the input into a specified range, the non-linear activation function is used to deal with the complexity of the input sets. There are many activation functions such as Sigmoid, which condenses the range between zero and one, and others such as Maxout, TLDR, Tanh, and ReLU. 
	
\begin{figure}[t!]
\centering
\includegraphics[width=0.42\textwidth]{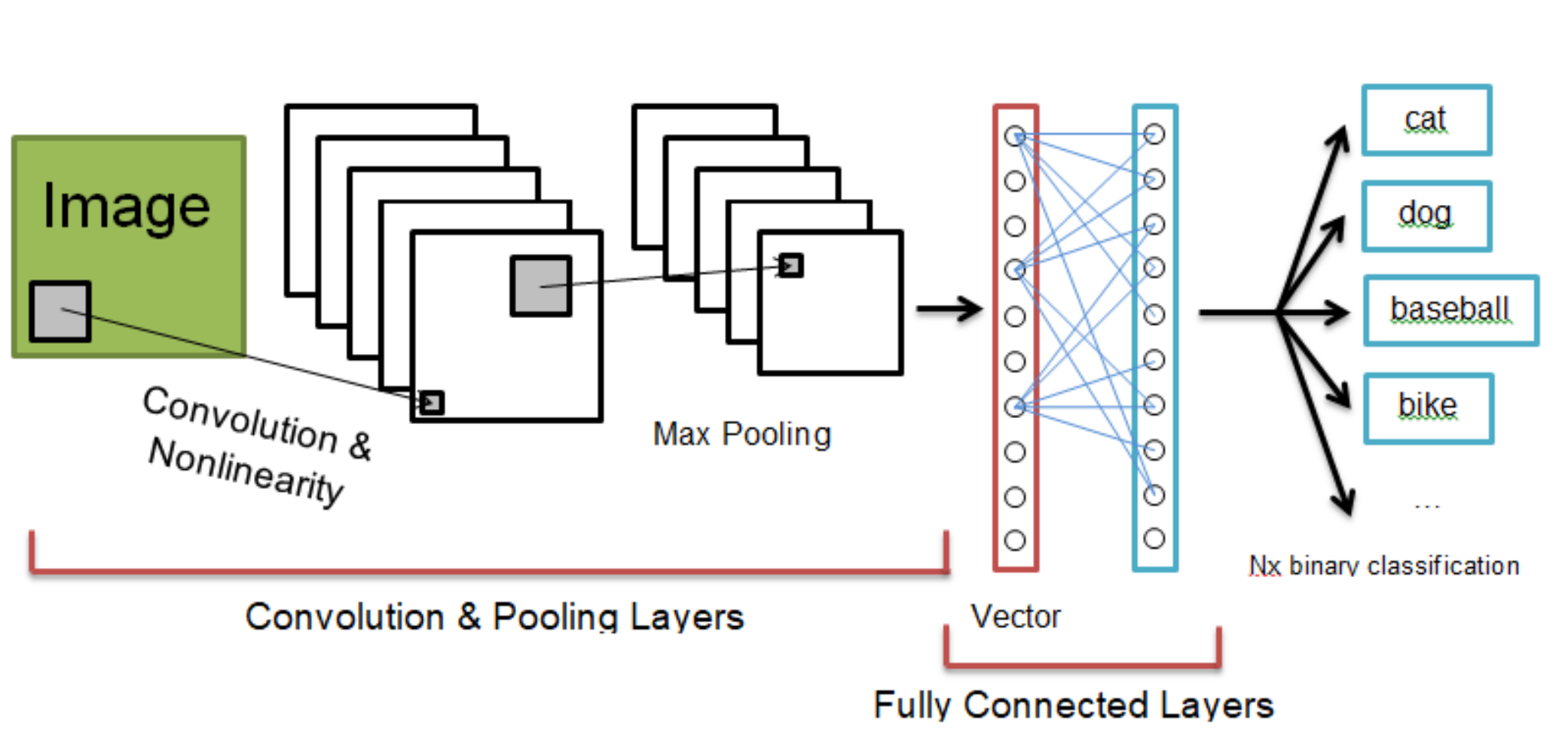}
\vspace*{-0.3cm}
\caption{ConvNet architecture for image classification applications (adopted from\cite{sze2017efficient}).}
\label{fig:ConvNet}
\end{figure}

%\subsection{Deep Neural Network}
	A Deep Neural Network (DNN) is simply a NN that has more than one hidden layer and is thus distinguished by its depth (i.e. the number of layers). Each layer of neurons in a DNN performs some kind of operation to train on a distinct set of features that are based on the previous layer's output.  %A brief overview on two common types of DNNs are presented as follows: 
%\item \textbf{Multilayer Perception (MLP)}:
%A Multilayer Perceptron (MLP) is basically a fully connected network where each node in a layer is connected to all nodes in the following layer. It utilize the non-linear activation function to classify the data \cite{zanaty2012support}. 
%https://medium.com/@datamonsters/artificial-neural-networks-for-natural-language-processing-part-1-64ca9ebfa3b2
	A Convolutional Neural Network (CNN) is a DNN type utilized for processing unique information through a hierarchy of layers and is prominent in a variety of applications such as image processing, %\cite{chen2017deeplab}, 
	sentence classification, % \cite{ kim2014convolutional},
	semantic parsing, %\cite{yih2015semantic}, 
	and speech recognition \cite{abdel2014convolutional}. There are many components that are contained in a CNN that assist with operating the convolutions. As illustrated in Fig. \ref{fig:ConvNet}, these include different types of layers such as \textit{convolutional}, \textit{fully connected}, and \textit{pooling} layers. 
	The input layer maintains different values based on the applications. For example, if a CNN is used for image classification, the input layer holds the raw pixel values of the image \cite{CNN_note}. The input layer feeds the \textbf{\textit{CONV layer}} which is represented by four sub-layers 1) \textit{convolution sub-layer} performs a dot product between the weights of the regionally-connected neurons and their input sets to compute the output, 2) \textit{non-linear sub-layer} uses a ReLU activation function to map the weight sum of regionally-connected neurons to $max(0,$ \textit{weight sum of regionally-connected neurons}$)$, 3) \textit{normalization sub-layer} scales the range of distributions of feature values to prevent the learning process to over compensate the correction in one weight dimension whereas under-compensating in another one \cite{stackoverflow}, 4) \textit{pool sub-layer} reduces the spatial size to make the number of parameters and computations to be less and less within the  network \cite{CNN_note}. The \textbf{\textit{Fully Connected (FC) layer}} forms the output of the previous \textbf{\textit{CONV layer}} as a vector that represents the list of feature values. Next, this vector can be converted to a stack of fully connected layers to identify the set of votes. The majority of votes determine the class scores.

\vspace{-0.2cm}

%\subsection{Training in DNN}
%A general goal regarding DNNs is making prediction using the trained network. This is known as \textbf{\textit{inference}} and it is only possible after a network has been trained. The process of training a DNN is that of learning a new capability from existing data and this is done by finding a set of weights and/or biases that minimize the average loss, the difference between the ideal correct output and the actual output, over a large training set; backpropagation is the key to training a DNN. It is an efficient algorithm for computing the gradient of the cost function. It is a simple implementation of chain rule of derivatives that provides the ability to compute all required partial derivatives in linear time.  Backprop computes the gradient of the error with respect to all weights in the network. \textit{Gradient Descent} uses the gradient to update all values, weights, and parameters such that the output error would be minimized. 	

\subsection{Binarized Neural Network (BNN)}
 In BNNs, the entire weights and activations are quantized with one- or two-bits with a small scarification in the classification accuracy \cite{courbariaux2016binarized}. Such novelty results in representing the BNNs associated parameters in a smaller memory footprint that can fit on an on-chip memory in the BNN accelerator.
\vspace*{-0.2cm}

\subsection{DNN Accelerator}
\vspace*{-0.2cm}
The advance of deep neural networks have stimulated the study of novel hardware architectures to fulfill the high demand for  computation power and memory bandwidth. For instance, billions of floating-point operations are required in a modern CNN to classify a single image \cite{umuroglu2017finn}. This not only requires the incorporation of customized computation-centric architecture, it also necessitates the removal of off-chip memory bottlenecks to maintain the DNN associated parameters on-chip. Furthermore, the inference operations in DNN must be executed in a fraction of second in the safety-critical applications such as autonomous vehicles \cite{azizimazreah2018tolerating}. This urges the researchers to explore a novel set of hardware optimizations to meet latency constraints. 
The conventional DNN accelerators are equipped with arrays of processing elements and multiple on-chip buffers. The processing elements enable concurrent execution of sparse dependence  multiply-accumulate (MAC) operations. The on-chip buffers store the input feature maps, weights, partial sums, and output feature maps. Despite the fact that the large-size  feature maps and weights might deprive DNN from the benefits of low access latency to on-chip storage, the temporal and spatial locality observed in the feature maps and weights enable us to deploy on-chip caching mechanism. In addition, a large on-chip storage is embedded in modern DNN accelerators to avoid the expensive traffic to access off-chip memory while maintaining the feature maps and weights near the processing elements \cite{azizimazreah2018tolerating}. Beside these, various techniques such as compression \cite{lin2018supporting}, pruning \cite{parashar2017scnn}, and reduced precision \cite{de2017understanding} have been devised in the past to improve the DNN accelerators' performance and to amortize their energy consumption overhead. Nevertheless, the current trends for designing DNN accelerators \textit{barely consider the essential requirement for maintaining the sophisticated processing elements and the on-chip buffers to be resilient in the presence of soft errors}.

In this study, we targeted a well-known category of DNN accelerators called BNN inference accelerator. BNN stores the weights and activations in 1-bit and 2-bit datatypes to significantly reduces the required memory for storing the network. This approach facilitates the dedication of \underline {exclusive on-chip memory} for maintaining the BNN information which results in significant off-chip memory access reduction.  

\section{Evaluation}

%\subsection{The Selected DNN Accelerator for Study}
% Even though this data representation has significantly reduced the costly off-chip memory accesses, but it also has increased the impact of transient faults on the results. In particular, condensing a 32-bit floating point number which was originally used to demonstrate weights and activations to few bits comes with the risk of loosing the whole synapse information if the representative BNN bit set flips due to soft error. The BNN accelerator that we used for our study, utilizes 1-bit and 2-bit for all input activations, weights and output activations that makes it highly vulnerable to soft errors. As it will be presented in the next section, if the soft error causes the data corruption that will be reused later in the dataflows of BNN accelerator, the contaminated data will pollute the remaining computations which results in the misclassification.  

\begin{comment}
\begin{figure}[t!]
	\centering
\includegraphics[width=0.4\textwidth]{Fig/CNN_Accelerator.pdf}
	\caption{CNN Accelerator (adopted from \cite{du2018reconfigurable})}
	\label{fig:CNN_accelerator_PE}
	\end{figure}
\end{comment}

\subsection{Experimental Setup}
In our experiments, we consider two different BNN topologies for evaluating fault injection: 
\begin{itemize}
    \item The convolutional network topology (\textbf{cnv}) determined in BNN is inspired by BinaryNet \cite{courbariaux2016binarized} and VGG-16 \cite{VGG_16} which is tailored with a 6 convolutional layers, 3 max pool layers and 3 fully connected layers. This topology is used to classify the CIFAR-10 dataset, which is categorized to two groups depends on the data representation for weight and activation: \textit{cnvW1A1} requires 1-bit to store the aforementioned parameters while the 2-bit data type is used in \textit{cnvW2A2}. There are around \underline{1.6 million} susceptible bits and \underline{3.2 million} susceptible bits to soft errors in W1A1 and W2A2 topologies, respectively. 
    
    \item The \textbf{LFC} is constructed by four fully connected layers tailored with 1024 neurons each layer. This network classifies the MNIST dataset. For such network topology, there are around \underline{3 million} susceptible bits to soft errors in both \textit{lfcW1A1} and \textit{lfcW1A2} networks.
\end{itemize}

It is noteworthy to indicate that the binarized algorithm is not applied on the inputs fed to the first layer and the outputs extracted from the last layer. Furthermore, since the fault injection is relatively long process for collecting meaningful data, we limited our experimental results on classifying 1000 images in CIFAR-10. This consideration reduces the time window required for classification. We uniformly distribute the faults across the memory/logic space and time to represent a realistic scenario of the effect of soft errors on BNN inference accelerator in each fault injection test.    

We run our experiments on a FPGA-processor co-design, Xilinx Zynq-7000 ARM/FPGA SoC, where A dual-core ARM Cortex-A9 processor and Xilinx 7-series FPGA logic are deployed. The FINN framework presented \cite{umuroglu2017finn} is used for BNN inference acceleration where a group of images are classified. This process was executed on Cortex A9 cores. FINN reads the images into the shared DRAM and launches the accelerator for classification. In order to synthesize the corresponding bitfile, we used Vivado HLS and Vivado. %In addition, the clock frequency of 200 MHz was set for both Vivado HLS and Vivado.  

\subsection{Fault Injection Scenarios on BNN Inference Accelerator}
The soft errors might strike either memory storage or combinational logic in the hardware accelerator that cause the precision of the results to fall behind the minimum allowable correctness. For instance, the classifier misprediction due to soft error in CNNs running on a hardware accelerator as illustrated in Fig. \ref{soft_error}, results in intolerable incident in the mission-critical applications. 

In this study, we assumed that the effect of soft error is uniformly distributed across space and time which is in line with the studies presented in \cite{dixit2011impact}. Thus, we selected a fault space with uniform distribution over the period of workload execution and throughout the random locations in the targeted units. 
 We not only investigated the impact of  Single-Event Upset (SEU), but also studied how the Multi-Bit Upset (MBU) affects the output of the accelerator. Our motivation to examine the impact of both SEU and MBU is the study in \cite{dixit2011impact} that shows even though the SEUs are still the dominance of transient faults, the ratio of MBUs has significantly increased over past years. The reason behind this relative shift is the technology scaling which delivers the transistors with reduced dimension. Hence, the radiation-induced transient faults with less energy are able to unbalance the critical charge of adjacent transistors which result in the stored bits of the adjacent cells to be inverted and to cause a glitch in the combinational logic \cite{ khoshavi2016bit}. In order to mimic the behavior of an MBU, we targeted a burst of bits in size of 8-bit for bit-flipping which is aligned with the study demonstrated in \cite{schirmeier2016efficient}. We disregard the faults that might occur in the combinational logic and the control logic units since they are less sensitive to soft errors \cite{seifert2012soft}. For the sake of simplicity, we assume that CPU, main memory, and the memory bus are resilient to soft errors. 

We examine the effect of soft errors on the network topology and the data type used to represent the weights and activations similar to the work presented in \cite{li2017understanding}. The wights and activations are stored on on-chip buffers and are significantly reused over the period of BNN algorithm execution. To be specific, we targeted the following parameters in the network topology:

\begin{itemize}
    \item \textbf{\textit{Weights}}:
    The BNN accelerator \underline{\textbf{exclusively}} performs \textit{inference operations} on a large volume of images for classification. Since the weights in BNN are set during the training session, any changes in the weights might lead the fluctuations in the classification accuracy for the rest of BNN execution on FPGA.

    \item \textbf{\textit{Activations}}:
    As mentioned before, the BNN that we target for our study has been already trained. Thus, the soft error in the activations of a pre-trained BNN manifest itself as variations in the accuracy of classification. Since the corrupted activations will be reused in the rest of the workload execution, we expect that the classification accuracy to drop.   
   
    \item \textbf{\textit{Layers}}: The convolutional network topology determined in BNN is inspired by BinaryNet \cite{courbariaux2016binarized} and VGG-16 \cite{VGG_16} which is tailored with a 6 convolutional layers, 3 max pool layers and 3 fully connected layers. Since each category of layers has different parameters, we examine the impact of soft errors in each category, separately. We did not run our experiments on \textit{lfc} network due to its simple network topology.
    
\end{itemize}
% In order to investigate the impact of soft errors on the data type, we run our fault injections on two categories of data representations: 1) The weights and activcations are represented with 1-bit (\textbf{W1A1}), 2) 2-bit is used for representing the weights and activations (\textbf{W2A2}).

We performed 2000 fault injections for  scenarios shown in Table \ref{tab:FI_results}, to collect a sufficient pool of samples for determining the vulnerability of different network topologies to soft errors in BNN inference accelerator. To be specific, the faults are injected on the targeted parameters located in a certain memory addresses which can be accessed through our in-house fault injection script. This process is an on-the-fly process meaning that the fault/faults is/are injected while FINN is running on Zynq-7000 SoC board. 
We evaluated the sensitivity of the accelerator to a range number of faults that might occur in the operational lifetime of the device. As listed in Table \ref{tab:FI_results}, we assessed the impact of injecting 1, 2, 5, 10, 20, 50, and 100 faults %during the operational lifetime of the accelerator 
to highlight the impact of accumulated soft errors which is fairly realistic in DNN inference accelerators. We initialized the network to the default after each fault injection test.

\subsection{Results}
%\begin{table*}[]
%\centering
%\fontsize{6}{7.2}\selectfont
We evaluated the vulnerability of BNN accelerator against soft errors through injecting 1, 2, 5, 10, 20, 50, and 100 accumulated faults during the operational lifetime of the accelerator at random time on random location in  \textit{cnvW1A1}, \textit{cnvW2A2}, \textit{lfcW1A1}, and \textit{lfcW1A2} under two well-known categories of soft errors, SEU and MBU, as listed in Table \ref{tab:FI_results}. Each fault injection scenario was repeated for 2000 rounds. The average of the classification accuracy degradation across fault injection tests was considered for each scenario. Based on our experimental results, we observed the following incidents:

%=================   Table ==============
\begin{table*}[]
\begin{center}
\caption{Fault injection impact of BNN classification accuracy*}
\fontsize{6}{7.2}\selectfont
\vspace*{-0.2cm}
\begin{tabular}{lccccccccc}
\hline
\multicolumn{1}{|c|}{}                                                                                      & \multicolumn{1}{c|}{}                                                                                            & \multicolumn{4}{c|}{\textbf{Weight}}                                                                                                                                                                                                                                                                                                                                                        & \multicolumn{4}{c|}{\textbf{Activation}}                                                                                                                                                                                                                                                                                                                                                    \\ \cline{3-10} 
\multicolumn{1}{|c|}{}                                                                                      & \multicolumn{1}{c|}{}                                                                                            & \multicolumn{2}{c|}{\textbf{SEU}}                                                                                                                                                            & \multicolumn{2}{c|}{\textbf{MBU}}                                                                                                                                                            & \multicolumn{2}{c|}{\textbf{SEU}}                                                                                                                                                            & \multicolumn{2}{c|}{\textbf{MBU}}                                                                                                                                                            \\ \cline{3-10} 
\multicolumn{1}{|c|}{\multirow{-3}{*}{\textbf{\begin{tabular}[c]{@{}c@{}}Network\\ Topology\end{tabular}}}} & \multicolumn{1}{c|}{\multirow{-3}{*}{\textbf{\begin{tabular}[c]{@{}c@{}}\# of injected \\ faults\end{tabular}}}} & \multicolumn{1}{c|}{\textbf{\begin{tabular}[c]{@{}c@{}}Accuracy \\ Reduction\end{tabular}}} & \multicolumn{1}{c|}{\textbf{\begin{tabular}[c]{@{}c@{}}Effective\\  Faults (\%)\end{tabular}}} & \multicolumn{1}{c|}{\textbf{\begin{tabular}[c]{@{}c@{}}Accuracy \\ Reduction\end{tabular}}} & \multicolumn{1}{c|}{\textbf{\begin{tabular}[c]{@{}c@{}}Effective\\  Faults (\%)\end{tabular}}} & \multicolumn{1}{c|}{\textbf{\begin{tabular}[c]{@{}c@{}}Accuracy \\ Reduction\end{tabular}}} & \multicolumn{1}{c|}{\textbf{\begin{tabular}[c]{@{}c@{}}Effective\\  Faults (\%)\end{tabular}}} & \multicolumn{1}{c|}{\textbf{\begin{tabular}[c]{@{}c@{}}Accuracy \\ Reduction\end{tabular}}} & \multicolumn{1}{c|}{\textbf{\begin{tabular}[c]{@{}c@{}}Effective\\  Faults (\%)\end{tabular}}} \\ \hline
\multicolumn{1}{|c|}{}                                                                                      & \multicolumn{1}{c|}{\textbf{1 Fault}}                                                                            & \multicolumn{1}{c|}{-0.00143 (-0.18\%)}                                                     & \multicolumn{1}{c|}{28.0}                                                                      & \multicolumn{1}{c|}{-0.00132 (-0.16\%)}                                                     & \multicolumn{1}{c|}{38.0}                                                                      & \multicolumn{1}{c|}{0.0001 (0.01\%)}                                                        & \multicolumn{1}{c|}{62.0}                                                                      & \multicolumn{1}{c|}{0.00056 (0.07\%)}                                                       & \multicolumn{1}{c|}{66.0}                                                                      \\ \cline{2-10} 
\multicolumn{1}{|c|}{}                                                                                      & \multicolumn{1}{c|}{\textbf{2 Faults}}                                                                           & \multicolumn{1}{c|}{0.00021 (0.03\%)}                                                       & \multicolumn{1}{c|}{39.0}                                                                      & \multicolumn{1}{c|}{-0.00131 (-0.16\%)}                                                     & \multicolumn{1}{c|}{59.0}                                                                      & \multicolumn{1}{c|}{-0.00034 (-0.04\%)}                                                     & \multicolumn{1}{c|}{77.0}                                                                      & \multicolumn{1}{c|}{0.00096 (0.12\%)}                                                       & \multicolumn{1}{c|}{75.0}                                                                      \\ \cline{2-10} 
\multicolumn{1}{|c|}{}                                                                                      & \multicolumn{1}{c|}{\textbf{5 Faults}}                                                                           & \multicolumn{1}{c|}{-0.00036 (-0.04\%)}                                                     & \multicolumn{1}{c|}{76.0}                                                                      & \multicolumn{1}{c|}{-0.00093 (-0.12\%)}                                                     & \multicolumn{1}{c|}{84.0}                                                                      & \multicolumn{1}{c|}{0.00159 (0.20\%)}                                                       & \multicolumn{1}{c|}{87.0}                                                                      & \multicolumn{1}{c|}{0.00224 (0.28\%)}                                                       & \multicolumn{1}{c|}{88.0}                                                                      \\ \cline{2-10} 
\multicolumn{1}{|c|}{}                                                                                      & \multicolumn{1}{c|}{\textbf{10 Fualts}}                                                                          & \multicolumn{1}{c|}{-0.00083 (-0.10\%)}                                                     & \multicolumn{1}{c|}{83.0}                                                                      & \multicolumn{1}{c|}{-0.00141 (-0.18\%)}                                                     & \multicolumn{1}{c|}{88.0}                                                                      & \multicolumn{1}{c|}{0.00345 (0.43\%)}                                                       & \multicolumn{1}{c|}{94.0}                                                                      & \multicolumn{1}{c|}{0.00482 (0.60\%)}                                                       & \multicolumn{1}{c|}{97.0}                                                                      \\ \cline{2-10} 
\multicolumn{1}{|c|}{}                                                                                      & \multicolumn{1}{c|}{\textbf{20 Faults}}                                                                          & \multicolumn{1}{c|}{0.00097 (-0.12\%)}                                                      & \multicolumn{1}{c|}{86.0}                                                                      & \multicolumn{1}{c|}{-0.00297 (-0.37\%)}                                                     & \multicolumn{1}{c|}{96.0}                                                                      & \multicolumn{1}{c|}{0.00762 (0.95\%)}                                                       & \multicolumn{1}{c|}{98.0}                                                                      & \multicolumn{1}{c|}{0.00883 (1.10\%)}                                                       & \multicolumn{1}{c|}{95.0}                                                                      \\ \cline{2-10} 
\multicolumn{1}{|c|}{}                                                                                      & \multicolumn{1}{c|}{\textbf{50 Faults}}                                                                          & \multicolumn{1}{c|}{-0.00219 (-0.27\%)}                                                     & \multicolumn{1}{c|}{94.0}                                                                      & \multicolumn{1}{c|}{-0.00346 (-0.43\%)}                                                     & \multicolumn{1}{c|}{98.0}                                                                      & \multicolumn{1}{c|}{\cellcolor[HTML]{FFFE65}0.01963 (2.44\%)}                               & \multicolumn{1}{c|}{\cellcolor[HTML]{FFFFFF}99.0}                                              & \multicolumn{1}{c|}{\cellcolor[HTML]{FFFE65}0.03740 (4.65\%)}                               & \multicolumn{1}{c|}{\cellcolor[HTML]{FFFFFF}100.0}                                             \\ \cline{2-10} 
\multicolumn{1}{|c|}{\multirow{-7}{*}{\textbf{cnvW1A1}}}                                                    & \multicolumn{1}{c|}{\textbf{100 Faults}}                                                                         & \multicolumn{1}{c|}{-0.00244 (-0.30\%)}                                                     & \multicolumn{1}{c|}{94.0}                                                                      & \multicolumn{1}{c|}{\cellcolor[HTML]{FFFFFF}-0.00361 (-0.45\%)}                             & \multicolumn{1}{c|}{95.0}                                                                      & \multicolumn{1}{c|}{\cellcolor[HTML]{FD6864}0.05270 (6.54\%)}                               & \multicolumn{1}{c|}{\cellcolor[HTML]{FFFFFF}100.0}                                             & \multicolumn{1}{c|}{\cellcolor[HTML]{FD6864}0.06787 (8.43\%)}                               & \multicolumn{1}{c|}{\cellcolor[HTML]{FFFFFF}100.0}                                             \\ \hline
\multicolumn{1}{c}{\textbf{}}                                                                               & \multicolumn{1}{l}{}                                                                                             & \multicolumn{1}{l}{}                                                                        &                                                                                                & \multicolumn{1}{l}{}                                                                        &                                                                                                & \multicolumn{1}{l}{}                                                                        &                                                                                                &                                                                                             & \multicolumn{1}{l}{}                                                                           \\ \hline
\multicolumn{1}{|l|}{}                                                                                      & \multicolumn{1}{c|}{\textbf{1 Fault}}                                                                            & \multicolumn{1}{c|}{-0.00032 (-0.04\%)}                                                     & \multicolumn{1}{c|}{22.0}                                                                      & \multicolumn{1}{c|}{-0.00162 (-0.19\%)}                                                     & \multicolumn{1}{c|}{21.0}                                                                      & \multicolumn{1}{c|}{-0.00148 (-0.17\%)}                                                     & \multicolumn{1}{c|}{29.0}                                                                      & \multicolumn{1}{c|}{-0.00058 (-0.07\%)}                                                     & \multicolumn{1}{c|}{38.0}                                                                      \\ \cline{2-10} 
\multicolumn{1}{|l|}{}                                                                                      & \multicolumn{1}{c|}{\textbf{2 Faults}}                                                                           & \multicolumn{1}{c|}{-0.00192 (-0.23\%)}                                                     & \multicolumn{1}{c|}{38.0}                                                                      & \multicolumn{1}{c|}{-0.00155 (-0.18\%)}                                                     & \multicolumn{1}{c|}{56.0}                                                                      & \multicolumn{1}{c|}{-0.00118 (-0.14\%)}                                                     & \multicolumn{1}{c|}{65.0}                                                                      & \multicolumn{1}{c|}{-0.00123 (-0.14\%)}                                                     & \multicolumn{1}{c|}{65.0}                                                                      \\ \cline{2-10} 
\multicolumn{1}{|l|}{}                                                                                      & \multicolumn{1}{c|}{\textbf{5 Faults}}                                                                           & \multicolumn{1}{c|}{-0.00193 (-0.23\%)}                                                     & \multicolumn{1}{c|}{74.0}                                                                      & \multicolumn{1}{c|}{-0.00207 (-0.24\%)}                                                     & \multicolumn{1}{c|}{85.0}                                                                      & \multicolumn{1}{c|}{-0.00216 (0.25\%)}                                                      & \multicolumn{1}{c|}{90.0}                                                                      & \multicolumn{1}{c|}{-0.00190 (-0.22\%)}                                                     & \multicolumn{1}{c|}{86.0}                                                                      \\ \cline{2-10} 
\multicolumn{1}{|l|}{}                                                                                      & \multicolumn{1}{c|}{\textbf{10 Faults}}                                                                          & \multicolumn{1}{c|}{-0.00266 (-0.31\%)}                                                     & \multicolumn{1}{c|}{85.0}                                                                      & \multicolumn{1}{c|}{-0.00315 (-0.37\%)}                                                     & \multicolumn{1}{c|}{88.0}                                                                      & \multicolumn{1}{c|}{-0.00085 (-0.10\%)}                                                     & \multicolumn{1}{c|}{91.0}                                                                      & \multicolumn{1}{c|}{-0.00064 (-0.08\%)}                                                     & \multicolumn{1}{c|}{90.0}                                                                      \\ \cline{2-10} 
\multicolumn{1}{|l|}{}                                                                                      & \multicolumn{1}{c|}{\textbf{20 Faults}}                                                                          & \multicolumn{1}{c|}{-0.00395 (-0.46\%)}                                                     & \multicolumn{1}{c|}{97.0}                                                                      & \multicolumn{1}{c|}{-0.00373 (-0.44\%)}                                                     & \multicolumn{1}{c|}{96.0}                                                                      & \multicolumn{1}{c|}{0.00054 (0.06\%)}                                                       & \multicolumn{1}{c|}{95.0}                                                                      & \multicolumn{1}{c|}{-0.00058 (0.07\%)}                                                      & \multicolumn{1}{c|}{91.0}                                                                      \\ \cline{2-10} 
\multicolumn{1}{|l|}{}                                                                                      & \multicolumn{1}{c|}{\textbf{50 Faults}}                                                                          & \multicolumn{1}{c|}{-0.00403 (-0.47\%)}                                                     & \multicolumn{1}{c|}{96.0}                                                                      & \multicolumn{1}{c|}{-0.00404 (-0.47\%)}                                                     & \multicolumn{1}{c|}{97.0}                                                                      & \multicolumn{1}{c|}{0.00193 (0.23\%)}                                                       & \multicolumn{1}{c|}{94.0}                                                                      & \multicolumn{1}{c|}{0.00232 (0.27\%)}                                                       & \multicolumn{1}{c|}{97.0}                                                                      \\ \cline{2-10} 
\multicolumn{1}{|l|}{\multirow{-7}{*}{\textbf{cnvW2A2}}}                                                    & \multicolumn{1}{c|}{\textbf{100 Faults}}                                                                         & \multicolumn{1}{c|}{-0.00489 (-0.57\%)}                                                     & \multicolumn{1}{c|}{97.0}                                                                      & \multicolumn{1}{c|}{\cellcolor[HTML]{FFFFFF}-0.00378 (-0.44\%)}                             & \multicolumn{1}{c|}{96.0}                                                                      & \multicolumn{1}{c|}{0.00711 (0.83\%)}                                                       & \multicolumn{1}{c|}{97.0}                                                                      & \multicolumn{1}{c|}{\cellcolor[HTML]{FFFFFF}0.00601 (0.70\%)}                               & \multicolumn{1}{c|}{92.0}                                                                      \\ \hline
                                                                                                            & \multicolumn{1}{l}{}                                                                                             & \multicolumn{1}{l}{}                                                                        & \multicolumn{1}{l}{}                                                                           & \multicolumn{1}{l}{}                                                                        & \multicolumn{1}{l}{}                                                                           & \multicolumn{1}{l}{}                                                                        & \multicolumn{1}{l}{}                                                                           & \multicolumn{1}{l}{}                                                                        & \multicolumn{1}{l}{}                                                                           \\ \hline
\multicolumn{1}{|l|}{}                                                                                      & \multicolumn{1}{c|}{\textbf{1 Fault}}                                                                            & \multicolumn{1}{c|}{0.00005 (0.006\%)}                                                      & \multicolumn{1}{c|}{18.0}                                                                      & \multicolumn{1}{c|}{0.00003 (0.003\%)}                                                      & \multicolumn{1}{c|}{16.0}                                                                      & \multicolumn{1}{c|}{0.00002 (0.002\%)}                                                      & \multicolumn{1}{c|}{46.0}                                                                      & \multicolumn{1}{c|}{0.00001 (0.001\%)}                                                      & \multicolumn{1}{c|}{51.0}                                                                      \\ \cline{2-10} 
\multicolumn{1}{|l|}{}                                                                                      & \multicolumn{1}{c|}{\textbf{10 Faults}}                                                                          & \multicolumn{1}{c|}{0.00003 (0.004\%)}                                                      & \multicolumn{1}{c|}{69.0}                                                                      & \multicolumn{1}{c|}{0.00004 (0.004\%)}                                                      & \multicolumn{1}{c|}{72.0}                                                                      & \multicolumn{1}{c|}{0.00061 (0.062\%)}                                                      & \multicolumn{1}{c|}{77.0}                                                                      & \multicolumn{1}{c|}{0.01199 (1.22\%)}                                                       & \multicolumn{1}{c|}{78.0}                                                                      \\ \cline{2-10} 
\multicolumn{1}{|l|}{\multirow{-3}{*}{\textbf{lfcW1A1}}}                                                    & \multicolumn{1}{c|}{\textbf{100 Faults}}                                                                         & \multicolumn{1}{c|}{0.00040 (0.04\%)}                                                       & \multicolumn{1}{c|}{93.0}                                                                      & \multicolumn{1}{c|}{0.00030 (0.03\%)}                                                       & \multicolumn{1}{c|}{89.0}                                                                      & \multicolumn{1}{c|}{\cellcolor[HTML]{FFCCC9}0.05058 (5.14\%)}                               & \multicolumn{1}{c|}{94.0}                                                                      & \multicolumn{1}{c|}{\cellcolor[HTML]{FFCCC9}0.06466 (6.57\%)}                               & \multicolumn{1}{c|}{96.0}                                                                      \\ \hline
\textbf{}                                                                                                   &                                                                                                                  &                                                                                             &                                                                                                &                                                                                             &                                                                                                &                                                                                             &                                                                                                &                                                                                             &                                                                                                \\ \hline
\multicolumn{1}{|l|}{}                                                                                      & \multicolumn{1}{c|}{\textbf{1 Fault}}                                                                            & \multicolumn{1}{c|}{-0.00014 (-0.01\%)}                                                     & \multicolumn{1}{c|}{20.0}                                                                      & \multicolumn{1}{c|}{-0.00012 (-0.01\%)}                                                     & \multicolumn{1}{c|}{18.0}                                                                      & \multicolumn{1}{c|}{-0.00013 (-0.01\%)}                                                     & \multicolumn{1}{c|}{10.0}                                                                      & \multicolumn{1}{c|}{-0.00014 (-0.01\%)}                                                     & \multicolumn{1}{c|}{17.0}                                                                      \\ \cline{2-10} 
\multicolumn{1}{|l|}{}                                                                                      & \multicolumn{1}{c|}{\textbf{10 Faults}}                                                                          & \multicolumn{1}{c|}{-0.00013 (-0.01\%)}                                                     & \multicolumn{1}{c|}{79.0}                                                                      & \multicolumn{1}{c|}{-0.00012 (-0.01\%)}                                                     & \multicolumn{1}{c|}{69.0}                                                                      & \multicolumn{1}{c|}{-0.00013 (-0.01\%)}                                                     & \multicolumn{1}{c|}{82.0}                                                                      & \multicolumn{1}{c|}{0.00037 (0.04\%)}                                                       & \multicolumn{1}{c|}{68.0}                                                                      \\ \cline{2-10} 
\multicolumn{1}{|l|}{\multirow{-3}{*}{\textbf{lfcW1A2}}}                                                    & \multicolumn{1}{c|}{\textbf{100 Faults}}                                                                         & \multicolumn{1}{c|}{0.00450 (0.46\%)}                                                       & \multicolumn{1}{c|}{88.0}                                                                      & \multicolumn{1}{c|}{0.00078 (0.08\%)}                                                       & \multicolumn{1}{c|}{86.0}                                                                      & \multicolumn{1}{c|}{0.00031 (0.03\%)}                                                       & \multicolumn{1}{c|}{87.0}                                                                      & \multicolumn{1}{c|}{0.00055 (0.06\%)}                                                       & \multicolumn{1}{c|}{88.0}                                                                      \\ \hline
                                                                  \\ \hline
\end{tabular}
\label{tab:FI_results}
\end{center}
\vspace*{-0.15cm}
\fontsize{6}{7.2}\selectfont
*These results are collected through performing 2000 fault injection tests for each scenario. In particular, we assessed the impact of injecting 1, 2, 5, 10, 20, 50, and 100 faults during the operational lifetime of the accelerator in \textit{cnvW1A1}, \textit{cnvW2A2}, \textit{lfcW1A1}, and \textit{lfcW1A2} to highlight the impact of accumulated soft errors. We only reported the effect of injecting 1, 10, and 100 faults on \textit{lfcW1A1} and \textit{lfcW1A2}. For instance, the last row of the table indicates that 100 faults are injected at random time on random locations while the classification algorithm is using \textit{lfcW1A2} network for classifying the images. The  \underline{Effective Runs} column indicates what percentage of test runs had faults that cause the overall accuracy to differ from the baseline.
\end{table*}

%===================================================
\begin{figure*}[t!]
\centering
\includegraphics[width=0.93\textwidth]{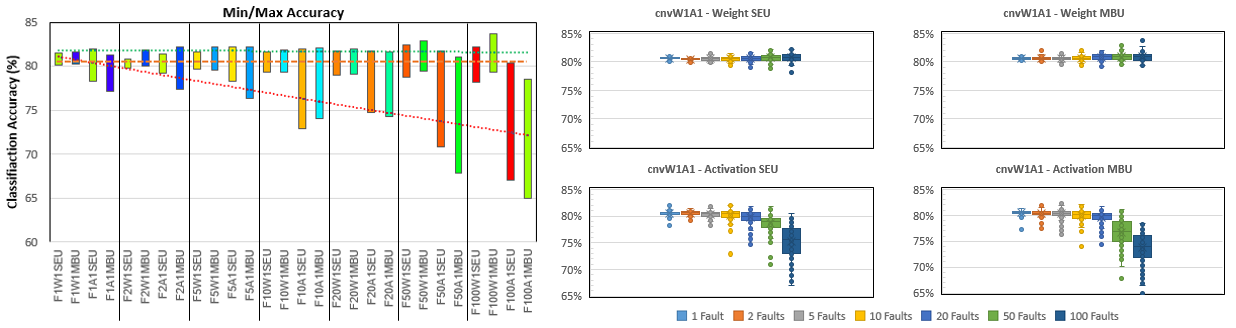}
\vspace*{-0.3cm}
\caption{The distribution of classification accuracy degradation in the presence of various faults in a \underline{cnvW1A1} network. The baseline classifies 1000 images with 80.5\% accuracy. The plot has been sectioned based on the number of injected faults. On the most left figure, the plot's bars that show the min/max classification accuracy due to the effect of SEUs on weights and activations are colored with the same color in each section. A different color is used for plotting MBUs' effects on weights and activations in each section.} 
\label{min_max_w1A1}
\end{figure*}

%===================================================
\begin{comment}
\begin{figure*}[t!]
\centering
\includegraphics[width=0.8\textwidth]{Fig/min_max_w2a2.PNG}
\vspace*{-0.3cm}
\caption{The min/max classification accuracy in the presence of various faults in a \underline{cnvW2A2} network. The baseline classified 1000 images with 85.3\% accuracy. The outliers and the accuracy distribution due to the effect of faults are illustrated with color-codded dots.}
\label{min_max_w2A2}
\end{figure*}
\end{comment}

%===================================================
\begin{figure*}[t!]
\centering
\includegraphics[width=0.96\textwidth]{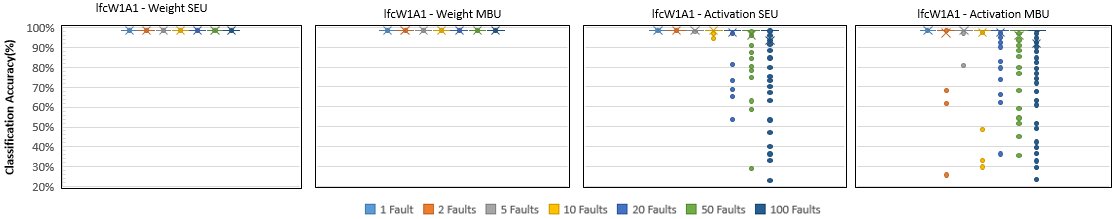}
\vspace*{-0.3cm}
\caption{The distribution of classification accuracy degradation in the presence of various faults in a \underline{lfcW1A1} network. The baseline classifies 1000 images with 98.4\% accuracy. The outliers and the accuracy distribution due to the effect of faults are illustrated with color-codded dots.}
\label{min_max_lfcw1a1}
\end{figure*}

%===================================================
\begin{figure*}[t!]
\centering
\includegraphics[width=0.96\textwidth]{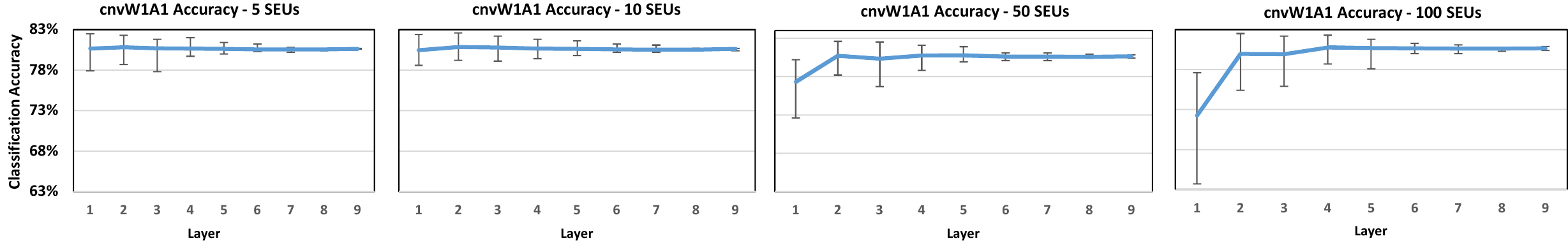}
\vspace*{-0.4cm}
\caption{The effect of different number of accumulated SEUs on different layers of \underline{cnvW1A1}.}
\label{layer_order_effect}
\end{figure*}

%===================================================

\begin{itemize}
    \item \textit{Reducing the number of bits for storing the network information increases the vulnerability of the accelerator to soft errors}. This effect has been highlighted with three different colors on \textit{Activation} column in Table \ref{tab:FI_results}. Even though the representation of weight and activation with one bit significantly reduces the memory size, it also drastically increases the negative effect of soft errors. As an example, the accuracy of the classification dropped by 8.43\% after injecting 100 MBUs in the activation layers of \textit{cnvW1A1} during the lifetime of the accelerator while our similar research conduction on \textit{cnvW2A2} caused less than 1\% accuracy degradation.
    \item \textit{The activation layer functions are significantly vulnerable to both SEUs and MBUs}. As shown in the \textit{Activation} column of Table \ref{tab:FI_results}, the effect of the faults that have directly bit flipped the bit set of the activation layers resulted in significant drop in the classification accuracy. This incident is more noticeable for the cases that the number of accumulated injected faults are more than 20 faults. We have color coded six cells in the \textit{Activation} column of Table \ref{tab:FI_results} to emphasize the drastic variations in the accuracy of classification for these incidents.
    \item \textit{As the number of accumulated faults increases in the fault injection test, the accuracy of the classification is relatively reduced}. For instance, the classification accuracy in a \textit{cnvW1A1} network can significantly drop from 80.5\% to 72.0\%, on average in the scenario that 100 accumulated faults are injected during the lifetime of BNN accelerator.
    \item \textit{The effect of MBU is relatively higher than SEU}. The MBU causes a set of bits to be flipped. This not only causes the variation in the targeted parameter, but it also causes the corruption of adjacent bits in other tensors which escalates the impact of the soft error. 
    \item \textit{Even though the average degradation of classification in all scenarios might not seem significant, we observed that the accelerator can potentially suffer from a \underline{drastic misclassification} in the worst case scenarios}. Fig. \ref{min_max_w1A1} and Fig. \ref{min_max_lfcw1a1} illustrates the variance of changes in the classification accuracy under different scenarios. Based on Fig. \ref{min_max_lfcw1a1} ({\fontfamily{cmtt}\selectfont lfcW1A1 - Activation SEU}), the \textit{lfcW1A1} network experiences a very wide range of misclassifications. For instance, the accuracy of image classifier can drastically drop by \textbf{76.7\%} (from 98.4\% to 22.92\%) in \textit{lfcW1A1}. This is the worst case scenario where 100 SEUs are injected during the workload operation.
    \item We conducted the fault injection tests on different layers of \textit{cnvW1A1} and \textit{cnvW2A2} with various accumulated faults. The results showed that \textit{the vulnerability of a layer appears to be directly related to how early it appears in the network, with the first layer being the most vulnerable by far} based on Fig. \ref{layer_order_effect}. 
\end{itemize}

\vspace*{-0.3cm}
\section{Related Works}

The authors of \cite{fault_inject2017} explored the potential of fault injection attacks on DNN to misclassify the given input. %The fault can modify the parameters in DNN which results in  misclassification for a particular input pattern. 
It assumes that  faults are systematically effective and are able to modify the parameters in the algorithm-level which lead to changes in the bias and the targeted layers in DNN. However, this approach lacks a realistic fault injection scenario. The transient faults or soft errors are uniformly distributed across space and time which makes the effect of them on the hardware unpredictable. Furthermore, the precise assumed memory faults injection attacks such as laser beam fault injection \cite{barenghi2012fault} and row hammer attack \cite{kim2014flipping} might not reproduce the same results in each attack due to changes in the parameters of the underlying device through the period of test \cite{KHOSHAVI201710, khoshavi2014applicability}. The authors of \cite{li2017understanding} used a DNN simulator to run the fault injections. They modified the framework written in C++ to realize the DNN hardware accelerator and to study the impact of fault on the underlying microarchitecture. Again, this approach does not consider a realistic framework for exploring the impact of faults on the DNN accelerator. On the other hand, the authors of \cite{azizimazreah2018tolerating} have proposed a circuit-level fault injector to mimic the effect of particle strikes on one or multiple nodes. Even though this approach can accurately simulates the faults' effect on the components of the circuit, but its high-latency and complexity to simulate architectural DNN accelerator makes it undesirable for the researchers who are studying the resiliency of DNN accelerators as an overarching system.      

\section{Conclusions}
In this paper, we showed that using the compression techniques for reducing the memory size of DNN has significantly increased the vulnerability of some parameters in the network to the soft errors. Our realistic FPGA-based fault injection method proved that the activation layer functions are significantly vulnerable to both SEUs and MBUs compared to the weight layers. We also demonstrated that the MBU has relatively higher impact on the accelerator. Furthermore, the soft errors have higher effect on the layers that appear earlier in the network. If the impact of accumulated soft errors in the accelerators are not decontaminated, it might result in drastic accuracy degradation in the output of the workloads.

\bibliographystyle{IEEEtran}
\fontsize{6}{7.2}\selectfont
\bibliography{biblio}
\end{document}